\documentclass[10pt,twocolumn,letterpaper]{article}

\usepackage{iccv}
\usepackage{times}
\usepackage{epsfig}
\usepackage{graphicx}
\usepackage{amsmath}
\usepackage{amssymb}

\usepackage{url}
\usepackage{enumitem}
\setlist[itemize]{topsep={0pt},partopsep={0pt}}
\usepackage{multirow} 
\usepackage{algorithm}
\usepackage{algpseudocode}
\usepackage{tikz}
\usepackage{comment}
\usepackage{color}

\definecolor{citecolor}{HTML}{1976D2}
\definecolor{linkcolor}{rgb}{0.956,0.298,0.235} 
\definecolor{fontcolor}{rgb}{0.267,0.420,0.809} 

\definecolor{gray}{gray}{0.95}
\definecolor{cyan}{rgb}{0.831,0.901,0.945}

\usepackage[pagebackref,breaklinks,colorlinks,linkcolor=linkcolor,citecolor=citecolor,bookmarks=false]{hyperref}
\usepackage[accsupp]{axessibility} 
\iccvfinalcopy % *** Uncomment this line for the final submission

\def\iccvPaperID{5613} % *** Enter the ICCV Paper ID here

\ificcvfinal\fi

\begin{document}

%%%%%%%%% TITLE
\title{Bootstrap Motion Forecasting With Self-Consistent Constraints}

% \author{First Author\\
% Institution1\\
% Institution1 address\\
% {\tt\small firstauthor@i1.org}
% % For a paper whose authors are all at the same institution,
% % omit the following lines up until the closing ``}''.
% % Additional authors and addresses can be added with ``\and'',
% % just like the second author.
% % To save space, use either the email address or home page, not both
% \and
% Second Author\\
% Institution2\\
% First line of institution2 address\\
% {\tt\small secondauthor@i2.org}
% }

\author{
Maosheng Ye$^1$\thanks{Work done during an internship at DeepRoute.AI.},
Jiamiao Xu$^2$, Xunnong Xu$^2$, Tengfei Wang$^1$, Tongyi Cao$^2$, Qifeng Chen$^1$
\\ [0.25cm]
$^1$The Hong Kong University of Science and Technology~~
$^2$DeepRoute.AI~~
}
\maketitle
% Remove page # from the first page of camera-ready.
\ificcvfinal\thispagestyle{empty}\fi

%%%%%%%%% ABSTRACT
\begin{abstract}
   We present a novel framework to bootstrap \textbf{M}otion forecast\textbf{I}ng with \textbf{S}elf-consistent \textbf{C}onstraints (MISC). The motion forecasting task aims at predicting future trajectories of vehicles by incorporating spatial and temporal information from the past. %The cross-modal domain property poses great challenges to ensure the stability of the trajectory prediction against small disturbances. 
A key design of MISC is the proposed Dual Consistency Constraints that regularize the predicted trajectories under spatial and temporal perturbation during training. Also, to model the multi-modality in motion forecasting,  we design a novel self-ensembling scheme to obtain accurate teacher targets to enforce the self-constraints with multi-modality supervision. With explicit constraints from multiple teacher targets, we observe a clear improvement in the prediction performance. Extensive experiments on the Argoverse motion forecasting benchmark and Waymo Open Motion dataset show that MISC significantly outperforms the state-of-the-art methods. As the proposed strategies are general and can be easily incorporated into other motion forecasting approaches, we also demonstrate that our proposed scheme consistently improves the prediction performance of several existing methods.
\end{abstract}

%%%%%%%%% BODY TEXT
\section{Introduction}

Motion forecasting has been a crucial task for self-driving vehicles that aims at predicting the future trajectories of agents (e.g., cars, pedestrians) involved in traffic. The predicted trajectories can further help self-driving vehicles plan their future actions and avoid potential accidents. Since the future is not deterministic, motion forecasting is intrinsically a multi-modal problem with substantial uncertainties. This implies that an ideal motion forecasting method should produce a distribution of future trajectories or at least multiple most likely ones. 

Due to the inherent uncertainty, motion forecasting remains challenging and unsolved yet. %Recently, several methods have been proposed to motion prediction by improving feature learning or optimizing pipeline.
% Rasterization based methods~\cite{bansal2018chauffeurnet,chai2019multipath} is one of the most direct ways to process the HDMap information, which converts the road elements into structural tensor format. With which, standard image backbones can be directly applied to extract the HDMap features. Works~\cite{gao2020vectornet,zhao2020tnt,gu2021densetnt} explore succinct representation for feature encoding. It applies graph neural networks on the vector representation to extract the features of road elements. LaneGCN~\cite{liang2020learning} furtherly models the lane topology as graph learning problem. The LaneConv operation focuses on the lane instances, which can better model the relationship between HDMap and traffic participants. LaneRCNN~\cite{zeng2021lanercnn} extends the framework of LaneConv and integrate with trajectory proposals from sampling approaches. Moreover, TPCN~\cite{ye2021tpcn} introduces the point cloud learning into motion forecasting task, which combines spatial and temporal learning in a unified framework to explore more compact representation. Some transformer based approaches~\cite{ngiam2021scene,liu2021multimodal} have been proposed to utilize the capability of transformer in modeling the long-term relationship across spatial and temporal dimension.
Recently, researchers have proposed different architectures based on various representations to encode the kinematic states and context information from HDMap in order to generate feasible multi-modal trajectories~\cite{bansal2018chauffeurnet,chai2019multipath,gao2020vectornet,gu2021densetnt,liang2020learning,liu2021multimodal,ngiam2021scene,varadarajan2021multipath++,ye2021tpcn,zeng2021lanercnn,zhao2020tnt}. These methods follow a traditional static training pipeline, where frames of each scenario are split into historical frames (input) and future frames (ground truth) in a fixed pattern. Nevertheless, the prediction task is a streaming task in real-world applications, where the current state will become a historical state as time goes by, and the buffer of the historical state is  a queue structure to make successive predicted trajectories. As a result, temporal consistency thus becomes a crucial requirement for the downstream tasks for fault and noise tolerance. To tackle this issue, trajectory stitching is widely applied in traditional planning algorithms~\cite{fan2018baidu} to ensure stability along the temporal horizon. However, as the trajectory stitching operation is non-differentiable, it cannot be easily incorporated into learning-based models. Though deep-learning-based models show unprecedented motion prediction performance compared with traditional counterparts,  they do not explicitly consider temporal consistency, leading to unstable behaviors in downstream tasks such as planning.

Inspired by these phenomena, we raise a question: can we explicitly enforce consistency when training a deep motion prediction model? On the one hand, the predicted trajectories should be consistent given the successive inputs along the temporal horizon, namely temporal consistency. On the other hand, the predicted trajectories should be stable and robust  against small spatial noise or disturbance, namely spatial consistency. In this work, we propose a self-supervised scheme, named as \textbf{\emph{MISC}}, to enforce consistency constraints in both spatial and temporal domains, namely \emph{Dual Consistency Constraints}. \emph{Dual Consistency Constraints} could be viewed as an inner-model consistency and can significantly improve  the quality and robustness of motion forecasting, without the need for extra data.  
% which introduces the consistency constraints in both spatial and temporal domains, namely Dual Consistency Constraints for developing invariant representations.

On top of the inner-model consistency, we also exploit the intra-model consistency. Multi-modality is another core characteristic of the motion prediction task. Existing datasets~\cite{chang2019argoverse,sun2020scalability} only provide a single ground-truth trajectory for each scenario, which can not satisfy multi-choice situations such as junction scenarios. Most methods adopt the winner-takes-all (WTA)~\cite{lee2016stochastic} or its variants~\cite{breuer2021quo,narayanan2021divide} to alleviate this situation. However, WTA tends to produce confused predictions when two trajectories are very close. In contrast, our method addresses the multi-modality problem by using more robust teacher targets obtained from self-ensembling, which leverages intra-model consistency. Multiple teacher targets can be viewed as a special kind of intra-model distillation while alleviating the problem of multi-modality.   Our contributions are summarized as follows,
\begin{itemize}[noitemsep,topsep=2pt] 
\item We propose self-consistent constraints in both intra and inner model aspects. 
\item For the inner-model consistency, Dual Consistency Constraints are proposed to enforce temporal and spatial consistency in our model, which is shown to be a general and effective way to improve the overall performance in motion forecasting.
\item For the intra-model consistency constraints, a self-ensembling constraint  is explicitly exploited  to enforce self-consistency with teacher targets, which provides multi-modality supervision for training.
\item Extensive experiments on the Argoverse~\cite{chang2019argoverse} motion forecasting benchmark and Waymo Open Motion dataset~\cite{ettinger2021large} show that the proposed approach achieves state-of-the-art performance.
\end{itemize}

\section{Related Work}
\noindent\textbf{Motion Forecasting.} Traditional methods~\cite{houenou2013vehicle,schulz2018interaction,xie2017vehicle,ziegler2014making} for motion forecasting mainly utilize HDMap information for the prior estimation and Kalman filter~\cite{kalman1960new} for motion states prediction. With the recent progress of deep learning on big data, more and more works have been proposed to exploit the potential of data mining in motion forecasting. Early efforts~\cite{bansal2018chauffeurnet,chai2019multipath,duvenaud2015convolutional,gao2020vectornet,henaff2015deep,liang2020learning,liu2021multimodal,shuman2013emerging,song2022learning,ye2021tpcn,zeng2021lanercnn,zhou2022hivt} explore different representations, including rasterized image, graph representation, point cloud representation and transformer to generate the features for the task and predict the final output trajectories by regression or post-processing sampling. Most of these works focus on finding more effective and compact ways of feature extraction on the surrounding environment (HDMap information) and agent interactions. Based on these representations, other approaches~\cite{casas2018intentnet,mangalam2020not,song2022learning,zeng2021lanercnn,zeng2019end,zhao2020tnt} try to incorporate the prior knowledge with traditional methods, which take the predefined candidate trajectories from sampling or clustering strategies as anchor trajectories. To some extent, these candidate trajectories can provide better guidance and goal coverage for the trajectories regression due to straightforward HDMap encoding. Nevertheless, this extra dependency makes the stability of models highly related to the quality of the trajectory proposals.
Goal-guided approaches~\cite{gilles2021home,gu2021densetnt,gilles2022gohome} are therefore introduced to optimize goals in an end-to-end manner, paired with sampling strategies that generate the final trajectory for better coverage.
% Therefore, some goal-guided works~\cite{gilles2021home,gu2021densetnt} optimize the goals in an end-to-end manner. This kind of method often adopts the sampling strategy to generate the final trajectory in order to achieve more coverage rate.

% Different from the above methods, our approach mainly aims to integrate consistency learning into prediction. Due to the unique property of this task which consists of spatial and temporal learning, it is natural to apply consistency constraints in both dimensions to enhance the robustness against disturbances.

\noindent\textbf{Consistency Regularization.} Consistency regularization has been fully studied in semi-supervised and self-supervised learning. Temporally related works~\cite{wang2019learning,lei2020blind,zhou2017unsupervised} apply pairwise matching to minimize the alignment difference through optical flow or correspondence matching to achieve temporal smoothness. Other works~\cite{bachman2014learning,foldiak1991learning,Ouyang_2021_ICCV,sajjadi2016regularization,Wang_2021_ICCV,simard1991tangent} apply consistency constraints to predictions from the same input with different transformations in order to obtain perturbation-invariant representations. 
 ~\cite{chakraborty2022improving, bhattacharyya2022ssl} reverse the temporal order or mask some information and generate pairwise consistency between these predicted trajectories.  ~\cite{su2022narrowing} introduced consistency by examining the gap between agent-centric and scene-centric settings.
% Our work can be seen as a combination of both types of consistency to fully consider the spatial and temporal continuity in motion forecasting.

% \noindent\textbf{Pseudo Label.} Pseudo label~\cite{lee2013pseudo} has been effective strategies for semi-supervised and unsupervised learning to obtain more labelled data. 

\noindent\textbf{Multi-hypothesis Learning.} Motion forecasting task inherently has multi-modality due to the future uncertainties and difficulties in acquiring accurate ground-truth labels. WTA~\cite{guzman2012multiple,sriram2019hierarchical} in multi-choice learning and its variants~\cite{makansi2019overcoming,rupprecht2017learning} incorporate with better distribution estimation to improve the training convergence, thus allowing more multi-modality. Some anchor-based methods~\cite{breuer2021quo,chai2019multipath,phan2020covernet,zeng2021lanercnn} introduce pre-defined anchors based on kinematics or road graph topology to provide guidance. However, these methods only allow one target per training stage. Other methods~\cite{breuer2021quo,gu2021densetnt} try to generate multi-target for supervision with heavy handcrafted optimizations. We propose a Teacher-Target-Constraints approach to provide more precise trajectory teacher labels by leveraging the power of self-ensembling~\cite{lee2013pseudo,zheng2021se}. Multiple targets are explicitly provided to each agent to better model the multi-modality.

\section{Approach}
The overall architecture of MISC comprises three parts. 1) We first utilize a joint spatial and temporal learning framework TPCN~\cite{ye2021tpcn} to extract pointwise features. Based on these features, we decouple the trajectory prediction problem as a two-stage regression task. The first stage performs goal prediction and completes the trajectory with the goal position guidance. The second stage takes the output of the first stage as anchor trajectories for refinement. 2) To enhance the spatial and temporal consistency of our MISC, we introduce \emph{Dual Consistency Constraints} at the inner-model level, which helps regularize the predictions in a streaming task view.
3) We leverage self-ensembling to generate more precise teacher targets to provide intra-model level self-consistent \textbf{Teacher Targets Constraints} in Sec.~\ref{sec:mpt_supervision}

\subsection{Architecture}
\label{sec:architecture}
% \vspace{-5px}
\begin{figure*}
    \centering
    \includegraphics[width=13cm]{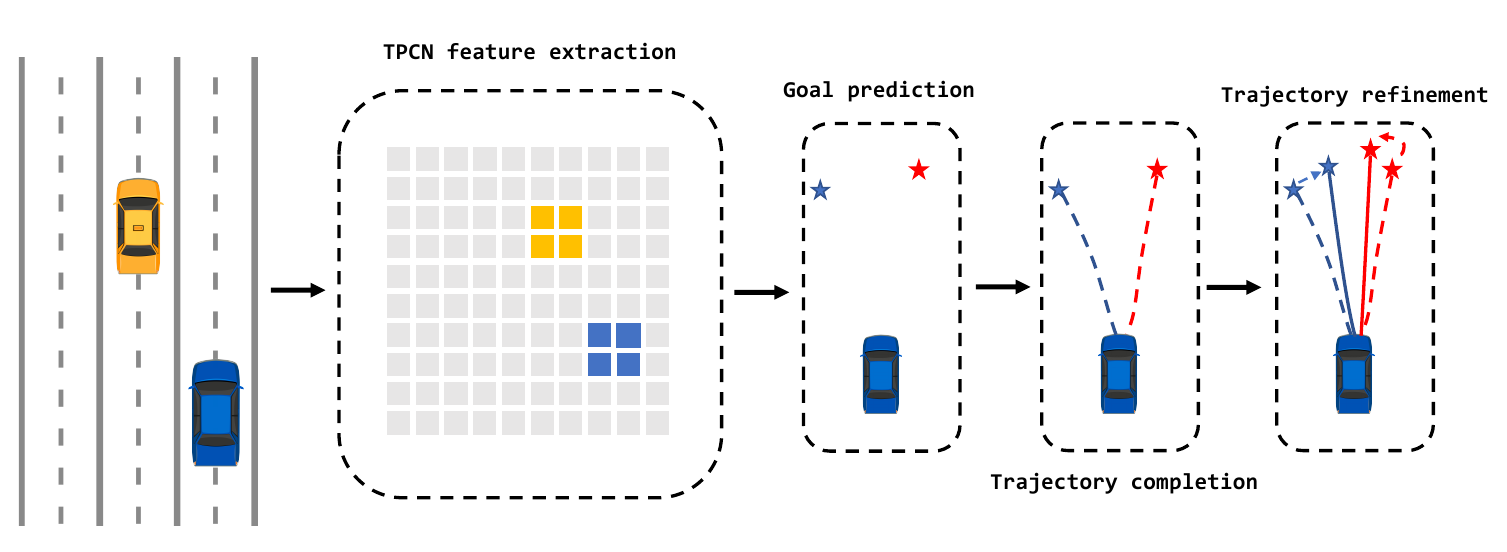}
    % \vspace{-10px}
    \caption{The overall architecture. We utilize TPCN~\cite{ye2021tpcn} as a feature extraction backbone to model the spatial and temporal relationship among agents and map information. A goal prediction header is then used to regress the possible goal candidates; with the goal position, we apply trajectory completion to obtain full trajectories; finally, the trajectories are refined based on the output of the trajectory completion module as anchor trajectories.}
    \label{fig:network}
\end{figure*}

Recently, TPCN~\cite{ye2021tpcn} has gained popularity in this task due to its flexibility for joint spatial-temporal learning and scalability to adopt more techniques from point cloud learning. Considering its limitation in representing future uncertainty, we extend TPCN in a two-stage manner through goal position prediction for more accurate waypoints prediction as our baseline. The pipeline is shown in Fig.~\ref{fig:network}. 

\textbf{Feature Extraction:} TPCN utilizes dual-representation point cloud learning techniques with multi-interval temporal learning to model the spatial and temporal relationship. All the historical trajectories of input agents  and map information are based on pointwise representation $\left\{ {{{\bf{p}}_{1}}}, {{{\bf{p}}_{2}}}, \hdots,  {{{\bf{p}}_{N}}} \right\}$, where 
${{\bf{p}}_{i}}$ is the $i$-th point with $N$ points in total, and then go through multi-representation learning framework to generate pointwise features $\mathcal{P} \in {R^{N \times C}}$, where $C$ is the channel number.

\textbf{Goal Prediction:} With the pointwise features from the backbone, we also adopt the popular goal-based ideas~\cite{gilles2021home,gu2021densetnt,zhao2020tnt} to find the optimal planning policy. Specifically, we first gather all corresponding pointwise agent features and then sum over features to get the agent instance feature $\phi \in {R^{1 \times C}}$. To generate $K$ goal position prediction $G = \{G^{k}: (g_x^{k}, g_y^{k})| 1 \leq k \leq K\}$, we use a simple MLP layer: $G = MLP(\phi)$. 
% Compared with previous goal-based methods, we do not rely on heavy sampling strategies and generate too many goal proposals, leading to a large computation overhead.
Instead of relying on heavy sampling strategies like previous goal-based methods, our method avoids generating extra proposals, which may lead to a large computation overhead.

\textbf{Trajectory Completion:} With the predicted goal positions, we need to complete each trajectory conditioned on these goals. We propose a simple trajectory completion module to generate $K$ full trajectories $\left\{\tau_{reg}^{k} | 1 \leq k \leq K\right\}$ with a single MLP layer as follows:
\begin{align}
&\tau_{reg}^{k} = \{(x_{1}^{k}, y_{1}^{k}), (x_{2}^{k}, y_{2}^{k}), \dots, (x_{T}^{k}, y_{T}^{k})\},\\
&\tau_{reg}^{k} =MLP(concat( \phi, G^{k})).
\end{align}
 
\textbf{Trajectory Refinement:} Inspired by Faster-RCNN~\cite{Ren2015Faster} and Cascade-RCNN~\cite{cai2018cascade}, we use the output trajectories from the Trajectory Completion as anchor trajectories to refine trajectories and predict the corresponding possibility of each trajectory. In particular, the input of the trajectory refinement module will be the whole trajectory with agent historical waypoints $\tau_{history}$. With a residual block followed by a linear layer $Reg$ and $Cls$ respectively, we regress the delta offset to the first stage outputs $\Delta_{\tau_{{reg}}}= Reg(\tau_{reg}, \tau_{history})$ and corresponding scores $\tau_{cls} =\left\{c^{k} | 1 \leq k \leq K\right\}$ respectively, where $\tau_{{cls}} = Cls(\tau_{reg}, \tau_{history})$. The final output trajectories will be $\tau_{reg^{'}} = \Delta_{\tau_{{reg}}} + \tau_{reg}$.
% \begin{eqnarray}
%     \Delta_{\tau_{{reg}}} &=& Reg(\tau_{reg}, \tau_{history}), \\
%     \tau_{{cls}} &=& Cls(\tau_{reg}, \tau_{history}).
% \end{eqnarray}

% \noindent The final output trajectories will be $\tau_{reg^{'}} = \Delta_{\tau_{{reg}}} + \tau_{reg}$.
% \begin{align}
% \tau_{reg^{'}} = \Delta_{\tau_{{reg}}} + \tau_{reg}.
% \end{align}

\subsection{Dual Consistency Constraints}
Consistency regularization has been proved as an effective self-constraint that improves robustness against disturbances. We thus propose inner-model level \textbf{Dual Consistency Constraints} in both spatial and temporal domains to align predicted trajectories for continuity and stability.
\vspace{-4mm}
\subsubsection{Temporal Consistency}
% \vspace{-5px}
\label{sec:temporal_consistency}
\begin{figure*}
    \centering
    \includegraphics[width=16cm]{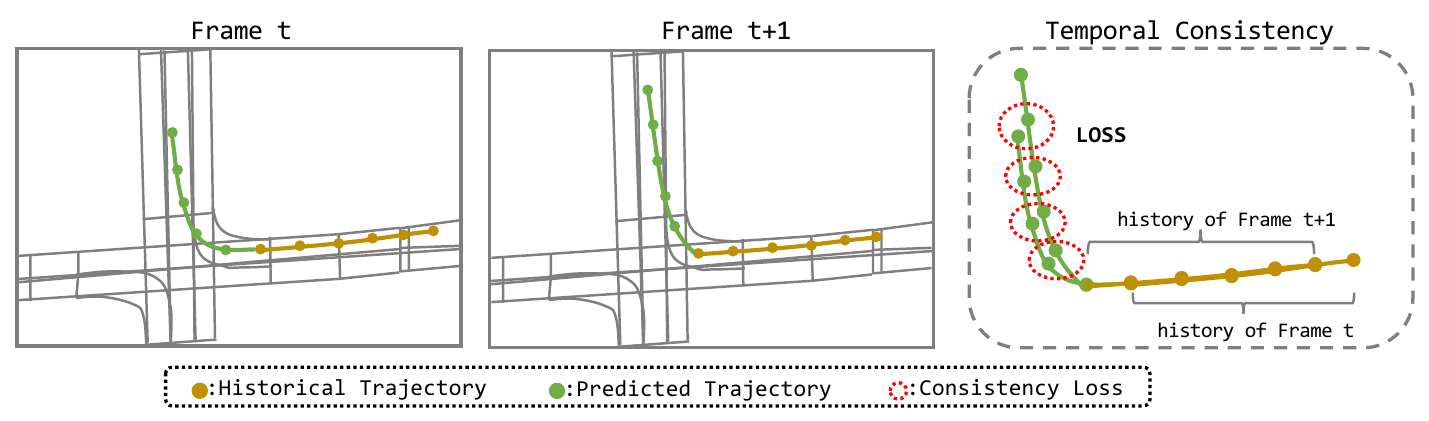}
    \caption{The overall idea of the temporal consistency. In the training stage, we first generate output prediction trajectory points as normal for each given scenario. Then we slide the input with a step in order to introduce the streaming nature to generate the consecutive output trajectory points. The proposed temporal consistency requires the overlap between these two outputs to be consistent}
    \label{fig:temporal_consistency}
    \vspace{-5px}
\end{figure*}
% Consistency Regularization has been fully studied in semi-supervised and self-supervised learning. Temporal related works~\cite{wang2019learning,lei2020blind,zhou2017unsupervised} have widely explored the idea for consistency in a cycle way. Most of the works apply pairwise matching and minimize the alignment difference through optical flow or correspondence matching to achieve the temporal smoothness. Some other works~\cite{bachman2014learning,foldiak1991learning,sajjadi2016regularization} apply consistency constraints to the predictions from the same input with different transformations in order to obtain perturbations invariant representations.
In motion forecasting, since each scenario contains multiple successive frames within a fixed temporal chunk,
% , any two sub-chunks as input data that have overlap should have consistent results. 
it is reasonable to assume that any two overlapping chunks of input data with a small time shift should produce consistent results. The motion forecasting task aims to predict $K$ possible trajectories with $T$ time steps for one scenario, given $M$ frames historical information. Suppose the information at each history frame is $I_i$, where $1 \leq i \leq M$ and the $k$-th output future trajectories are $\left\{(x^{k}_i, y^{k}_i) | M < i \leq M + T\right\}$. We first apply time step shift $s$ for the input for temporal consistency. Therefore, the input history frames information will be $\left\{I_i | 1 + s \leq i \leq M + s\right\}$ and then we apply the same network for the shifted history information with surrounding HDMap information to generate the $k$-th output trajectories $\left\{({x'^{k}_{i}}, y'^{k}_{i}) | M + s < i \leq M + s + T\right\}$. When $s$ is small, the driving intentions or behavior keeps stable for a short period. Since both trajectories have $T-s$ overlapping waypoints, they should be as close as possible and share consensus. Thus, we can construct self-constraints for a single scenario input due to the streaming property of the input data. Fig.~\ref{fig:temporal_consistency} demonstrates the overall idea of the temporal consistency constraint.

%Todo(yms) add to discussion
% Similar to knowledge distillation, the shifted input can be exposed to more future information including more map guidance, which could serve as the teacher network.

\noindent\textbf{Trajectory Matching:} Since we predict $K$ future trajectories to deal with the multi-modality, it is crucial to consider the trajectory matching relationship between original predictions and time-shifted predictions when applying the temporal consistency alignment. For a matching problem, the metric on similarity criteria and matching strategies will be two key factors. Several ways can be used to measure the difference between trajectories, such as Average Displacement Error (\textbf{ADE}) and Final Displacement Error (\textbf{FDE}). We utilize \textbf{FDE} as the criteria since the last position error can partially reflect the similarity with less bias from averaging compared with \textbf{ADE}.

\noindent\textbf{Matching Strategy:} There are roughly four ways used for matching, namely forward matching, backward matching, bidirectional matching, and Hungarian matching. Forward matching takes one trajectory in the current frame and finds its corresponding trajectory in the next frame with the least cost or maximum similarity. Backward matching is the reverse way compared to forward matching. Furtherly, bidirectional matching consists of both forward and backward matching, which considers the dual relationship. Hungarian matching is a linear optimal matching solution based on linear assignment. Forward and backward matching only considers the one-way situation, which is sensitive to noise and unstable. Hungarian matching has a high requirement for cost function choice. Based on these observations, we choose bidirectional matching as our strategy. We also show its advantages over the other approaches in Sec.~\ref{sec:ablation_study}.  After obtaining the optimal matching pairs $\left\{(m_k, n_k) | 1 \leq k \leq  K\right\}$, we can compute the consistency constraint by a simple smooth $L_{1}$ loss~\cite{Ren2015Faster} $\mathcal{L}_{Huber}$:
\begin{align}
    \mathcal{L}_\textnormal{temp} = \sum_{k=1}^{K}\sum_{t=s+1}^{T}\mathcal{L}_\textnormal{Huber}((x_{t}^{m_k}, y_{t}^{m_k}), (x{'}_{t-s}^{n_k}, y{'}_{t-s}^{n_k})).
\end{align}

\vspace{-3mm}
\subsubsection{Spatial Consistency}
Since our MISC is a two-stage framework, the second stage mainly aims for trajectory refinement. It will be more convenient to add spatial permutation in the second stage with less computational cost. First, we apply spatial permutation function $Z$, including flipping and random noise, to the trajectories from the first stage. The refinement module will process these augmented inputs and generate the offset to the ground truth and classification scores. Under the small spatial permutation and disturbance, we assume that the outputs of the network should also be self-consistent, meaning that the outputs have strong stability or tolerance to noise. Compared with data augmentation, it is explicit regularization. Then the spatial consistency constraint $\mathcal{L}_{spa}$ is as follows:
\begin{align}
\vspace{-10px}
    \mathcal{L}_\textnormal{spa} = \mathcal{L}_\textnormal{Huber}(\Delta_{\tau_{{reg}}}, Z^{-1}(Reg(Z(\tau_{reg}, \tau_\textnormal{history}))).
    \vspace{-10px}
\end{align}
% Since we process the input sequence by sequence in the trajectory refinement module, we do not need to care about the matching problem. It is exactly a one-on-one matching problem. Therefore, spatial consistency is more like an inner-input problem, while temporal consistency is an intra-input problem. Then the spatial consistency constraint $\mathcal{L}_{spa}$ is as follows: 

Then the total loss for Dual Consistency Constraints module will be $\mathcal{L}_{cons} = \mathcal{L}_{spa} + \mathcal{L}_{temp}$.
% \begin{align}
% \vspace{-10px}
%     \mathcal{L}_{cons} = \mathcal{L}_{spa} + \mathcal{L}_{temp}.
%     \vspace{-10px}
% \end{align}

\subsection{Teacher-Target Constraints}
\label{sec:mpt_supervision}
Teacher-Target Constraints enforce intra-model consistency that not only leverages the power of knowledge distillation but also helps alleviate the multi-modality supervision problem. Existing datasets~\cite{chang2019argoverse,sun2020scalability} only provide a single ground-truth trajectory for the target agent, which is to be predicted in one scenario. In order to encourage the multi-modality of models, the winner-takes-all (WTA) strategy is commonly used to prevent the model from collapsing into a single domain. However, the WTA training strategy suffers from instability associated with network initialization. Some other approaches~\cite{breuer2021quo,narayanan2021divide} introduce robust estimation methods to select better hypotheses. 
To some extent, these methods can only implicitly model the multi-modality. Some other approaches~\cite{breuer2021quo,zhao2020tnt} generate several possible future trajectories based on the kinematics model and road graph topology. DenseTNT~\cite{gu2021densetnt} only uses teacher labels to predict goal sets through a hill-climbing algorithm. These optimization methods tend to impose strict constraints and handcrafted prior knowledge, resulting in inaccurate teacher-targets and inferior performance. In contrast, our approach aims to generate more accurate teacher targets to provide explicit multi-modality supervision through self-ensembling to leverage the power of semi-supervised learning and knowledge distillation.
% ~\cite{tarvainen2017mean,yu2019uncertainty,zheng2021se}.

\begin{figure*}
    \centering
    % \vspace{-5px}
    \includegraphics[width=12cm]{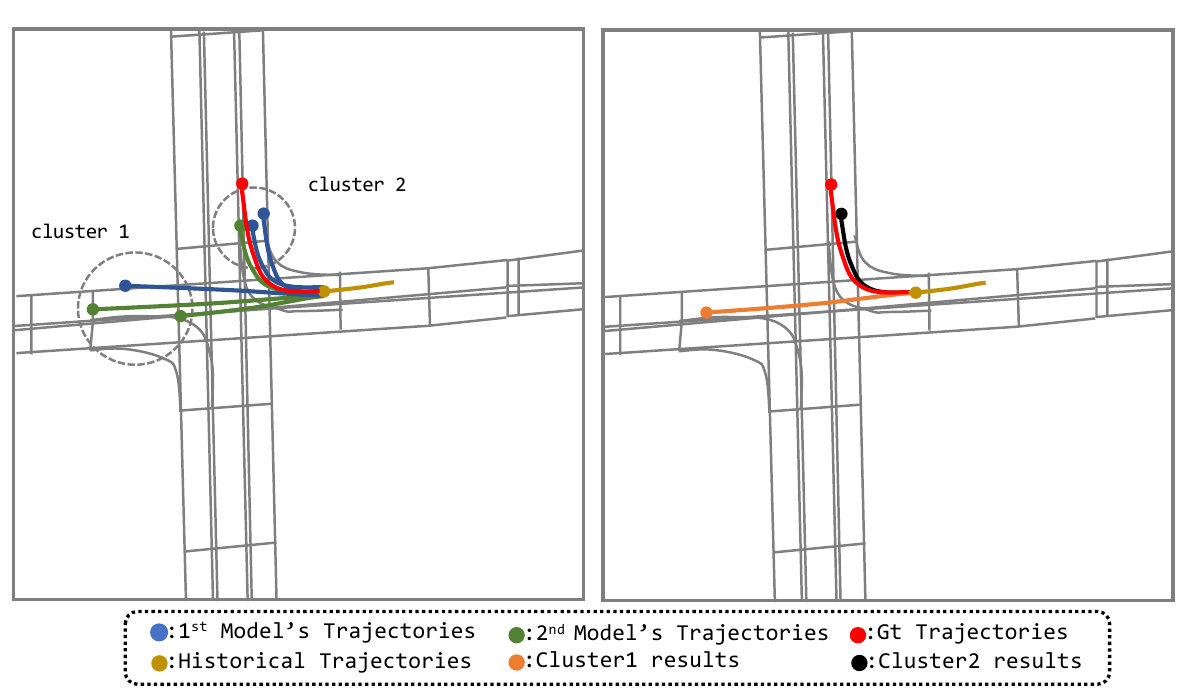}
    \caption{The overall procedure for the teacher-target generation. We obtain multiple predictions from outputs of different models for the target agents in each scenario; then we apply the K-means clustering algorithm to ensemble the trajectories}
    \label{fig:pseduo_target_generation}
    
\end{figure*}

\noindent\textbf{Teacher-Target Generation.} 
The key part of our approach lies in generating more accurate teacher labels for each agent. 
% It sounds like a ``chicken and egg'' problem since we need Teacher-Target to improve the performance. 
However, it is straightforward to apply model ensembling techniques~\cite{he2020structure,laine2016temporal,tarvainen2017mean} to obtain more powerful predictions. Compared with previous works~\cite{breuer2021quo,chai2019multipath,zhao2020tnt}, we do not rely on handcrafted anchor trajectory sampling, which is based on inaccurate prior knowledge, including motion estimation. Meanwhile, soft targets from ensembling can better finetune the predictions and reduce the gradient variance for better training convergence. As suggested in works~\cite{dietterich2000ensemble,opitz1999popular}, the prediction error decreases when the ensemble approach is used once the model is diverse enough. Therefore, we apply k-means algorithm~\cite{macqueen1967some} to the predicted trajectories that are collected within different training procedures (for example, launched with different seeds of random number generators, optimized with different learning rates, etc.) of MISC without Teacher-Target Constraints to generate $J$ trajectories with corresponding scores for each scenario.
Fig.~\ref{fig:pseduo_target_generation} shows the overall process of our approach. Then with the original ground-truth label, we will formulate $J + 1$ target trajectories as follows:
\vspace{-3mm}
\begin{align}
   &\tau_{conf} = \{c_{0}, c_{1}, \dots, c_{J}\}, \\
% \end{align}
% \begin{align}
   &\tau_{tgt}^{j} = \{(x_{1}^{tgt_{j}}, y_{1}^{tgt_{j}}), (x_{2}^{tgt_{j}}, y_{2}^{tgt_{j}}), \dots, (x_{T}^{tgt_{j}}, y_{T}^{tgt_{j}})\},
\end{align} 
where $\tau_{tgt}^{j}$ is the $j$-th trajectory with score $c^j$, among $J+1$ target trajectories. To simplify the notation, $\tau_{tgt}^{0}$ is the ground-truth trajectory with $c_{0}$ set to $1$.

% \noindent\textbf{Teacher-Target Matching.} Since we have a set of future trajectories and predictions for each target agent, the matching between future trajectories and predictions will be crucial. DETR~\cite{carion2020end} introduces the set predictor and utilizes Hungarian matching to pair the set of predictions and ground-truth labels. Instead of performing Hungarian matching, a naive approach can take the WTA strategy for each possible target. For each target, we find the predicted trajectory with the highest similarity to the source trajectory. The supervision will be enforced between the source and target trajectories.

\subsection{Learning}
The total supervision of our MISC can be decoupled into several parts, as described in previous sections. For the regression and classification parts, we loop over all the possible $J + 1$ targets $\tau_{tgt}$. For each target $\tau_{tgt}^{j}$ with confidence $\tau_{conf}^{j}$, we apply WTA strategy as described in Sec.~\ref{sec:mpt_supervision}. Suppose $k^{*}$-th trajectory from trajectory refinement output $\tau_{reg^{'}}$ is the best trajectory which has the maximum similarity with target $\tau_{tgt}^{j}$, the classification loss and regression loss are defined as:
\begin{align}
&\mathcal{L}^{j}_{cls} = \frac{1}{K}\sum_{k=1}^{K}\tau_{conf}^{j}\mathcal{L}_{Huber}(c^{k}, c^{k^{*}}),\\
  & \mathcal{L}^{j}_{reg} = \frac{1}{T}\sum_{t=1}^{T}\tau_{conf}^{j} \mathcal{L}_{Huber}((x_{t}^{k^{*}},y_{t}^{k^{*}}), (x_{t}^{tgt_{j}}, y_{t}^{tgt_{j}})).
%   \mathcal{L}^{j}_{reg} = \mathcal{L}^{j}_{reg} + \frac{1}{T}\sum_{i=1}^{T}\tau_{conf}^{j} f(x_{i}^{k^{*}} + x_{i}^{j} -\Delta{x_{i}^{k^{*}}} , y_{i}^{k^{*}} + y_{i}^{j}-\Delta{y_{i}^{k^{*}}}),
\end{align}
For classification loss design, we adopt the displacement prediction idea from TPCN~\cite{ye2021tpcn} to alleviate the hard assignment phenomenon. As for converting the displacement into probability, we use the standard softmin function to distribute the scores. Since we have trajectory completion and refinement modules, the regression loss will be $\mathcal{L}_{reg} = \sum_{j=0}^{J}(\mathcal{L}^{j}_{reg} + \mathcal{L}^{j}_{\Delta{reg}})$, where $\mathcal{L}^{j}_{\Delta{reg}}$ is the regression loss for the refinement module. The final loss is $\mathcal{L} = \mathcal{L}_{reg} + \mathcal{L}_{cls} + \mathcal{L}_{cons}$.

% \begin{equation}
%   \mathcal{L} = \mathcal{L}_{reg} + \mathcal{L}_{cls} + \mathcal{L}_{cons}
% \end{equation}

\section{Experiments}

% We conduct experiments on the Argoverse dataset~\cite{chang2019argoverse}, one of the largest publicly available motion forecasting datasets. We first compare our MISC with other state-of-the-art methods, and experiments show significant improvements in the evaluation metrics. Then we show the impact of dual consistency constraints on predicted trajectories with some qualitative results. Furthermore, we provide ablation studies to evaluate the effectiveness of each proposed module and design experiments for some hyperparameter choices. Finally, we extend our proposed modules to other methods to demonstrate their generalization ability.
We conduct experiments on the Argoverse dataset~\cite{chang2019argoverse}, one of the largest publicly available motion forecasting datasets. We compare our MISC with other state-of-the-art methods. Furthermore, we provide ablation studies to evaluate the effectiveness and generalization ability of each proposed module and design experiments for some hyperparameter choices.

\subsection{Experimental Setup}
\label{sec:exp_setup}
%TODO(YMS) REFINE
\noindent\textbf{Dataset.} Argoverse~\cite{chang2019argoverse}  provides more than 300K scenarios with rich HDMap information. For each scenario, objects are divided into three types: agent, AV and others, where  ``agent" is the object to be predicted. Moreover, each scenario contains $50$ frames sampled at $10$ Hz, meaning that the time interval between successive frames is $0.1$s. 
% Given the first $20$ frames in the scenario as historical context, the task is to predict future trajectories for the ``agent" objects in the next $30$ frames. 
The whole dataset is split into training, validation, and test sets, with $205,942$, $39,472$, and $78,143$ sequences, respectively. 
 Waymo open motion dataset (WOD) contains multiple types of agents including vehicles, pedestrians, and cyclists. A total of more than 100,000 segments are provided with more than 1500 km of roadway coverage.
% The test set only provides trajectories of the first $2$ seconds, with the ground-truth labels hidden. Furthermore, the map information which consists of lane points or polygons can be constructed by the given map API. 

\noindent\textbf{Metrics.} 
We use the standard evaluation metrics, including ADE and FDE. ADE is defined as the average displacement error between ground-truth trajectories and predicted trajectories over all time steps. FDE is defined as displacement error between ground-truth trajectories and predicted trajectories at the last time step. 
% Since the motion forecasting task has natural multi-modality, 
We predict $K$ candidate trajectories for each scenario and calculate the metrics with the ground truth labels. Accordingly, minADE and minFDE are minimum ADE and FDE over the top $K$ predictions. Moreover, miss rate (MR) is also considered, defined as the percentage of the best-predicted trajectories whose FDE is within a threshold ($2$m). Brier-minFDE is the minFDE plus $(1-p)^2$, where $p$ is the corresponding trajectory probability. Metrics for $K = 1$ and $K = 6$ are used in our experiments. Note that Brier-minFDE$_{6}$ is the ranking metric.

\noindent\textbf{Experimental Details.} We apply some data augmentation, including random flipping with a probability of $0.5$ and global random scaling with a scaling ratio between $[0.8, 1.25]$ during the training stage. As for model settings, the time shift $s$ for the temporal consistency constraint is set to $1$. We adopt $K=6$ to generate $6$ trajectories and use $J=6$ teacher targets for each scenario. Furthermore, we choose bidirectional matching for temporal consistency constraint. We finally use $10$ models for ensembling due to computation resource limits. For more training details, we have included them in the supplementary materials. 

\begin{table*}[!t]

\begin{center}
   \footnotesize
   \centering
   \def\arraystretch{1.4}
   \begin{tabular}{c|c c c|c c c c}
      \hline
      {Models} & {minADE$_{1}$} & {minFDE$_{1}$} & {MR$_{1}$} & {minADE$_{6}$} & {minFDE$_{6}$} & {MR$_{6}$} & {b-FDE$_{6}$}
      \\
    %   \specialrule{1pt}{0pt}{0pt}
      \hline
    %   Argoverse Baseline~\cite{chang2019argoverse} & 2.96 & 6.81 & 0.81 & 2.34 & 5.44 & 0.69
    %   \\
    %   \hline
    %   Argoverse Baseline\ (NN)~\cite{chang2019argoverse} & 3.45 &7.88 & 0.87 & 1.71 & 3.29 & 0.54
    %   \\
    %   \specialrule{1pt}{0pt}{0pt}
      Jean~\cite{chang2019argoverse,mercat2020multi} & 	1.74 & 4.24 & 0.68& 0.98 &1.42 &  0.13 & 2.12\\ 
    %   \hline
    %   uulm-mrm~\cite{chang2019argoverse} & 1.90& 4.19& 0.63  &0.94 &1.55 &0.22 &2.24  \\ 
      \hline
      LaneConv~\cite{liang2020learning} & 1.71 & 3.78 & 0.59 & 0.87 &  1.36 & 0.16 & 2.05
      \\
      \hline
      LaneRCNN~\cite{zeng2021lanercnn} & 1.68 & 3.69& 0.57 & 0.90 &  1.45 & 0.12 & 2.15
      \\
      \hline
      mmTransformer~\cite{liu2021multimodal} & 1.77 & 4.00 & 0.62 & 0.87 & 1.34 & 0.15 & 2.03
      \\
      \hline
      SceneTransformer~\cite{ngiam2021scene} & 1.81 & 4.06 & 0.59 & 0.80 & 1.23 & 0.126 & 1.88
      \\
      \hline
      TNT~\cite{zhao2020tnt} & 1.77 & 3.91 & 0.59 & 0.94 & 1.54 &  0.13 & 2.14
      \\
      \hline
      DenseTNT~\cite{gu2021densetnt} & 1.68 & 3.63 & 0.58 & 0.88 & 1.28 &  0.125 & 1.97
      \\
      \hline
      PRIME~\cite{song2022learning} & 1.91 & 3.82 & 0.59 & 1.22 & 1.55 &  0.12 & 2.09
      \\
      \hline
      TPCN~\cite{ye2021tpcn} & 1.58 & 3.49 & 0.56 & 0.88 & 1.24 &  0.13 & 1.92
      \\
      \hline
      HOME~\cite{gilles2021home} & 1.70 & 3.68 & 0.57 & 0.89 & 1.29 &  \bf 0.08 & 1.86
      \\
      \hline
      MultiPath++~\cite{varadarajan2021multipath++} & 1.623 & 3.614 & 0.564 & 0.790 &  1.214 &  0.13 & 1.793
      \\
      \hline
      Hivt++~\cite{zhou2022hivt} & 1.56 & 3.44 & 0.563 & 0.767 &  1.146 &  0.12 & 1.817
      \\
      
    %   \specialrule{1pt}{0pt}{0pt}
      \hline
      \def\arraystretch{1.2}
      Ours & \bf 1.476 & \bf 3.251 & \bf 0.532 & \bf 0.766 & \bf 1.135 & 0.11 & \bf 1.756
      \\
      
      \hline
   \end{tabular}
   
\end{center}
\caption{The detailed results of our MISC and other top-performing approaches on the Argoverse test set. And b-FDE$_{6}$ is the abbreviation of brier-minFDE$_{6}$}

\label{test_set_result}
\end{table*}
%todo(yms) change part?
\subsection{Experimental Results}
\subsubsection{Results on Argoverse Dataset}
\noindent\textbf{Argoverse Leaderboard Results.} We provide detailed quantitative results of our MISC on the Argoverse test set as well as public state-of-the-art methods in Tab.~\ref{test_set_result}. Compared with previous methods, our MISC improves all the evaluation metrics except MR$_6$ by a large margin. Furtherly, since the proposed modules are all general training components, other existing motion forecasting  models can also benefit greatly from these strategies. 
% Before the ECCV submission deadline, we achieve $\textbf{1st}$ place on the leaderboard for all $K=1$ evaluation metrics and outperform the other state-of-the-art approaches by a large margin. Meanwhile, we also rank $\textbf{1st}$ place for the main ranking metric brier-minFDE$_6$.

\begin{figure*}
    \centering
    \includegraphics[width=16cm]{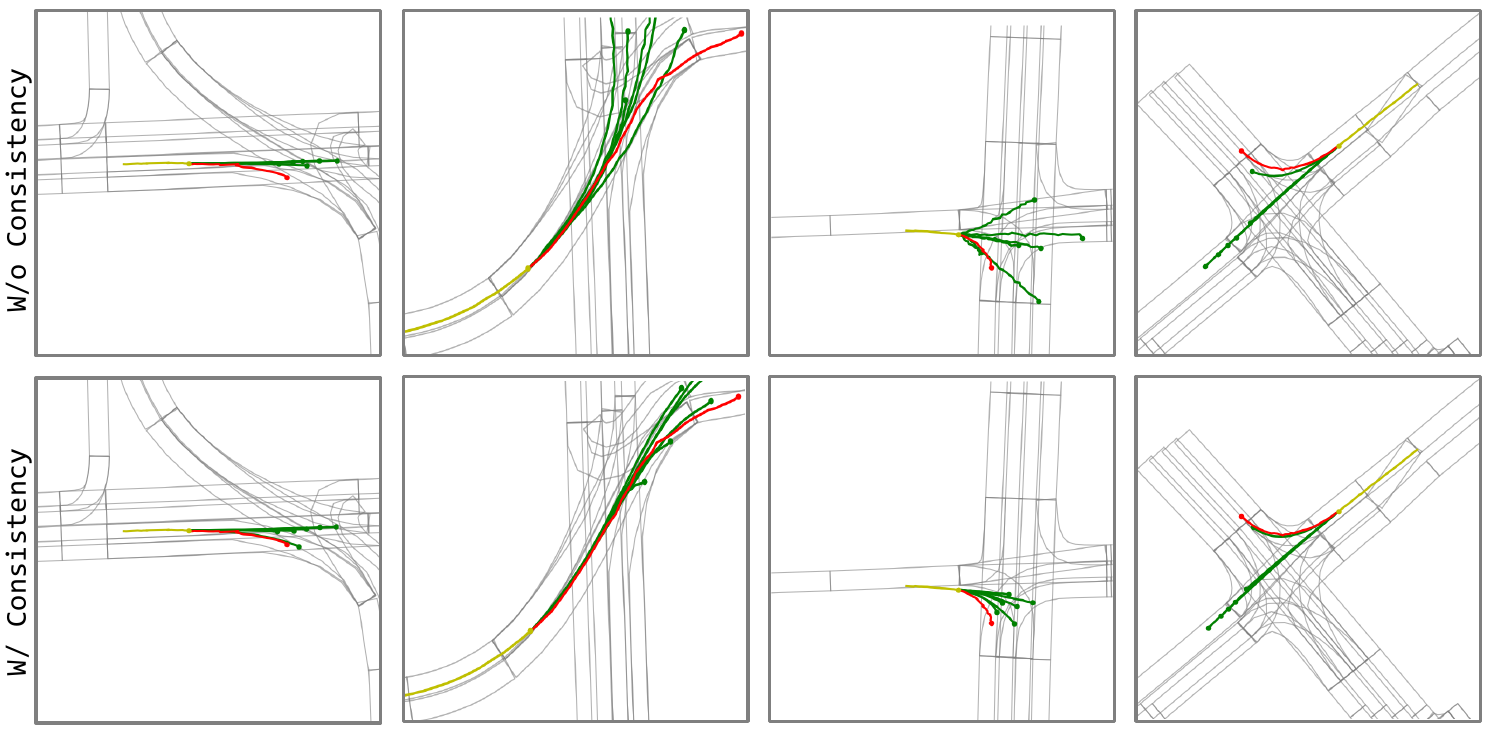}
    % \vspace{-5px}
    \caption{The past trajectory is in yellow, the predicted trajectory is in green, and the ground truth is in red. The top row of the figure shows the results without consistency, while the bottom row shows the results with consistency}
    \label{fig:qualitative_results}
    
\end{figure*}
\noindent\textbf{Qualitative Results.} We also present some qualitative results on the Argoverse validation set in Fig.~\ref{fig:qualitative_results}. Compared with results without consistency, the Dual Consistency Constraints improve both the quality and smoothness of the predicted trajectories significantly, resulting in more feasible and stable results despite the input noise.  

\subsubsection{Results on Waymo Open Motion Dataset}
We provide some quantitative results on the validation set of the Waymo Open dataset motion prediction task~\cite{ettinger2021large}, shown in Tab.~\ref{tab:waymo_results}. Compared with KEMP~\cite{lu2022kemp} and SceneTransformer~\cite{ngiam2021scene}, we also achieve very promising results and show comparable improvement, demonstrating the effectiveness of our approach. We also provide some ablation studies on WOD in the supplementary materials.

\begin{table}
% \vspace{-5px}
   \begin{center}
      \small
      \centering
      \setlength{\tabcolsep}{4pt}
      \begin{tabular}{cccccc}
      \hline 
      Method & minADE$\downarrow$ & minFDE$\downarrow$ & Miss Rate$\downarrow$ & mAP$\uparrow$\\
      \hline\hline
      Baseline~\cite{ettinger2021large} & 0.675 &  1.349 & 0.183 & 0.268 \\
      \hline
      KEMP~\cite{lu2022kemp} & 0.5691 & 1.1993& 0.1458& 0.394 \\
      \hline
     SceneTrans~\cite{ngiam2021scene}
 & 0.613 & 
1.22 &
0.157 & 
0.284 \\
      
      \hline
      Ours & \textbf{0.54} & \textbf{1.11} & \textbf{0.128} & \textbf{0.41} \\
      \hline
      
      \end{tabular}
   \end{center}
%   \vspace{-10px}

% \vspace{-10px}
\caption{Quantitative results on the validation set of the Waymo Open dataset motion prediction task.}
\label{tab:waymo_results}
\end{table}

% \vspace{-10px}
\subsection{Ablation Studies}
\label{sec:ablation_study}
\begin{table}
	\begin{center}
        \scriptsize
	\setlength\tabcolsep{3pt}
		\begin{tabular}{cc|cc|c|cc|cc}
			\hline
			\multicolumn{2}{c|}{Architecture} & \multicolumn{2}{c|}{Consistency} & \multirow{2}* {TTC} & \multicolumn{2}{c|}{K=1} & \multicolumn{2}{c}{K=6}\\
% 			\cline{1-3}
			Goal & Ref. & Temp. & Spatial &  & minADE & minFDE & minADE & minFDE\\
			\hline
			& & & & & 1.34 & 2.95 & 0.73 & 1.15\\
			\hline
			\checkmark& & & & & 1.33 & 2.91 & 0.725 & 1.10\\
			\hline
			\checkmark&\checkmark & & & & 1.31 & 2.89 & 0.71 & 1.07\\
			\hline
			\checkmark&\checkmark & \checkmark& & & 1.24 & 2.70 & 0.662 & 0.981\\
			\hline
			\checkmark&\checkmark &\checkmark &\checkmark & & 1.22 & 2.67 & 0.653 & 0.954 \\
			\hline
			\checkmark&\checkmark & & &\checkmark & {1.26} & {2.77} & {0.69} & {1.01} \\
			
			\hline
			\checkmark&\checkmark &\checkmark &\checkmark &\checkmark & \textbf{1.19} & \textbf{2.60} & \textbf{0.640} & \textbf{0.929} \\
			\hline
		\end{tabular}
	\end{center}
 \caption{Ablation study results of modules. Goal refers to Trajectory completion with goal prediction. ``Ref.'' is the trajectory refinement module, and the ``Temp.'' is temporal consistency. TTC refers to Teacher-Target Constraints during training}
	 \label{tab:module_ablation}
\end{table}

\begin{table}
	
	\begin{center}
 \resizebox{\linewidth}{!}{
 \scriptsize
	\setlength\tabcolsep{3pt}
		\begin{tabular}{@{}c|c|ccc|ccc@{}}
			\hline
			\multirow{2}*{Matching Strategy} &
			\multirow{2}*{Similarity} &
			\multicolumn{3}{c|}{K=1} & \multicolumn{3}{c}{K=6}\\
% 			\cline{1-3}
			 & & minADE & minFDE & MR & minADE & minFDE & MR\\
			\hline 
			 \multirow{2}{*}{Forward} & ADE & 1.25 & 2.70 & 0.46 & 0.670 & 0.982 & 0.089 \\ 
			 & FDE & 1.24 & 2.69 & 0.46 & 0.668 & 0.980 & 0.088\\ 
			 \hline
			 \multirow{2}{*}{Backward} & ADE & 1.25 & 2.70 & 0.46 & 0.670 & 0.982 & 0.089 \\ 
			 & FDE & 1.24 & 2.68 & 0.46 & 0.667 & 0.958 & 0.085\\ 
			 \hline
		     \multirow{2}{*}{Bidirectional} & ADE &\textbf{1.22} &\textbf{2.67} & 0.446& 0.666 & 0.972 & 0.087 \\ 
			 & FDE & \textbf{1.22} &\textbf{2.67} & \textbf{0.445} & \textbf{0.653} & \textbf{0.954}&\textbf{0.084} \\ 
			 \hline
			 \multirow{2}{*}{Hungarian} & ADE & 1.24 & 2.69 & 0.46 & 0.668 & 0.975 & 0.088 \\ 
			 & FDE & 1.23 & 2.69 & 0.45 &0.660 & 0.968 & 0.088\\ 
			 \hline
			
		\end{tabular}}
	\end{center}
        \caption{Ablation study on matching factor for temporal consistency. In this experiment, we remove the Teacher-Target Constraints to fairly study the effect}
	\label{tab:matching}
\end{table}

\begin{table}
	
	\begin{center}
  \scriptsize
	\setlength\tabcolsep{3pt}
		\begin{tabular}{{c|ccc|ccc}}
			\hline
			{Teacher Target Num}  & \multicolumn{3}{c|}{K=1} & \multicolumn{3}{c}{K=6}\\
% 			\cline{1-3}
			 $J$ & minADE & minFDE &MR & minADE & minFDE & MR\\
			\hline 
			 1 & {1.29} &{2.82} & {0.50} & {0.70} & {1.03}& {0.104}
			 \\
			\hline
			3  & {1.28} & {2.80} & {0.48} & \textbf{0.69} & {1.02}& {0.10}\\
			\hline
			6  & \textbf{1.26} & \textbf{2.77} & \textbf{0.47} & \textbf{0.69} & \textbf{1.01}& \textbf{0.09}\\
			\hline
		\end{tabular}
	\end{center}
 \caption{Ablation study results on the teacher target number $J$}
	\label{tab:Pseudo_Target_Num}
	
\end{table}

\begin{table}%[htbp]
% \vspace{-5px}
	\begin{center}
  \scriptsize
  \resizebox{\linewidth}{!}{
	\setlength\tabcolsep{3pt}
		\begin{tabular}{c|cc|cc|cc}
			\hline
			\multirow{2}* {Method} &
			\multirow{2}* {Consistency} & \multirow{2}* {TTC} &
			\multicolumn{2}{c|}{K=1} & \multicolumn{2}{c}{K=6}\\
% 			\cline{1-3}
			&  & & minADE & minFDE & minADE & minFDE\\
			
			\hline
			\multirow{3}{*}{LaneGCN~\cite{liang2020learning}} &$\times$&$\times$ & 1.35&2.97 &0.71 &1.08 \\
			&\checkmark & $\times$ &\textbf{1.29} &\textbf{2.80} &\textbf{0.68} & \textbf{1.00} \\
			&$\times$ &\checkmark  &\textbf{1.30}  &\textbf{2.88} &\textbf{0.69} & \textbf{1.04} \\
			\hline
			\multirow{3}{*}{TPCN~\cite{ye2021tpcn}} &$\times$&$\times$ & 1.34&2.95 &0.73 &1.15 \\
			&\checkmark &$\times$ &\textbf{1.27} &\textbf{2.79} &\textbf{0.69} & \textbf{1.04} \\
			&$\times$ &\checkmark  &\textbf{1.30} &\textbf{2.86} &\textbf{0.69} & \textbf{1.09} \\
			\hline
			%TODO(need to refine)%
			\multirow{3}{*}{mmTransformer~\cite{liu2021multimodal}} &$\times$&$\times$ & 1.38&3.03 &0.71 &1.15 \\
			&\checkmark & $\times$ &\textbf{1.31} &\textbf{2.83} &\textbf{0.68} & \textbf{1.02} \\
			& $\times$ &\checkmark  &\textbf{1.29} &\textbf{2.80} &\textbf{0.68} & \textbf{1.04}\\
			\hline
			\multirow{3}{*}{DenseTNT~\cite{gu2021densetnt}} &$\times$&$\times$ & 1.36&2.94 &0.73 &1.05 \\
			&\checkmark& $\times$&\textbf{1.25} &\textbf{2.81} &\textbf{0.68} & \textbf{0.98} \\
			& $\times$ &\checkmark  &\textbf{1.30} &\textbf{2.82} &\textbf{0.69} & \textbf{1.00} \\
			\hline
		\end{tabular}
  }
	\end{center}
% 	\vspace{-10px}

\caption{Ablation study of consistency constraints and Teacher Target Constraints on different state-of-the-art methods on Argoverse validation set. Performance for methods without constraints is obtained from corresponding papers or our reproduction
% Performance for methods without consistency constraints is obtained from corresponding papers, except the results of DenseTNT, which are based on our reproduction
}
\label{tab:consistency_study}
\end{table}

\noindent\textbf{Component Study.} As shown in Tab.~\ref{tab:module_ablation}, we conduct an ablation study for our MISC on the Argoverse validation set to evaluate the effectiveness of each proposed component. We adopt TPCN~\cite{ye2021tpcn} as the baseline shown in the first row of Tab.~\ref{tab:module_ablation} and add the proposed components progressively. The architecture modifications from the goal set prediction and trajectory refinement module show their promising improvements of about $2$\%. Dual consistency Constraints have the largest improvements of more than $5$\% among all the evaluation metrics. Especially for minFDE$_{1}$, temporal consistency can optimize $20$ cm, indicating the temporal constraints can improve both final position and trajectory probability prediction. Compared with temporal consistency, spatial consistency has less effect on models since we only enforce this constraint in the trajectory refinement stage. Finally, the Teacher-Target Constraints significantly increase performance, manifesting its effectiveness in helping training convergence.

% \begin{table*}[t]
% \vspace{-5px}
% \caption{Ablation study results of time-shift $s$ used by temporal consistency}
% 	\begin{center}
% 	\setlength\tabcolsep{3pt}
% 		\begin{tabular}{{c|ccc|ccc}}
% 			\hline
% 			{Time shift}  & \multicolumn{3}{c|}{K=1} & \multicolumn{3}{c}{K=6}\\
% % 			\cline{1-3}
% 			$s$  & minADE & minFDE &MR & minADE & minFDE & MR\\
% 			\hline 
% 			 1 & \textbf{1.22} &\textbf{2.67} & \textbf{0.444} & \textbf{0.653} & \textbf{0.954}& {0.084}
% 			 \\
% 			 \hline
% 			 2  & {1.23} &\textbf{2.67} & \textbf{0.444} & {0.654} & {0.958}&\textbf{0.082}\\
% 			\hline
% 			3  & {1.25} & {2.69} & {0.445} & {0.662} & {0.964}& {0.085}\\
% 			\hline
% 			4  & {1.25} & {2.70} & {0.446} & {0.667} & {0.969}& {0.086}\\
% 			\hline
% 		\end{tabular}
% 	\end{center}
	
% 	\label{tab:time_shift}
% 	\vspace{-10px}
% \end{table*}

%Todo(yms) consistency constraints or loss
\noindent\textbf{Temporal Consistency Factors.} 
% Since our consistency constraints module is a general training strategy, we also apply it to different popular models with state-of-the-art performance to show its generalization capability. As shown in Tab.~\ref{tab:consistency_study}, our consistency constraints can effectively improve the performance of models regardless of their representations through the training phase. There is a noticeable improvement of over $5\%$ on every metric, demonstrating the effectiveness of our Dual Consistency Constraints. 
% We conduct experiments on the hyperparameter $s$ that is used in the temporal consistency. 
We study the factors in the matching problems, including similarity and matching strategies. As shown in Tab.~\ref{tab:matching}, both Hungarian and Bidirectional matching show their advantages over the single direction matching. Although Hungarian matching can ensure the one-to-one matching relationship, it is sensitive to the similarity metric and numerical precision, both of which are not stable in the early training stage. In contrast, bidirectional matching with the FDE similarity metric nearly achieves the best results across all the evaluation metrics. Meanwhile, we also conduct experiments to find the best time-shift value $s$ in the temporal consistency. The details can be found in the supplement.
% Therefore, we choose bidirectional matching with the FDE similarity metric.
% through its bidirectional matching guarantee.

\begin{figure}
    \centering
    \includegraphics[width=8cm]{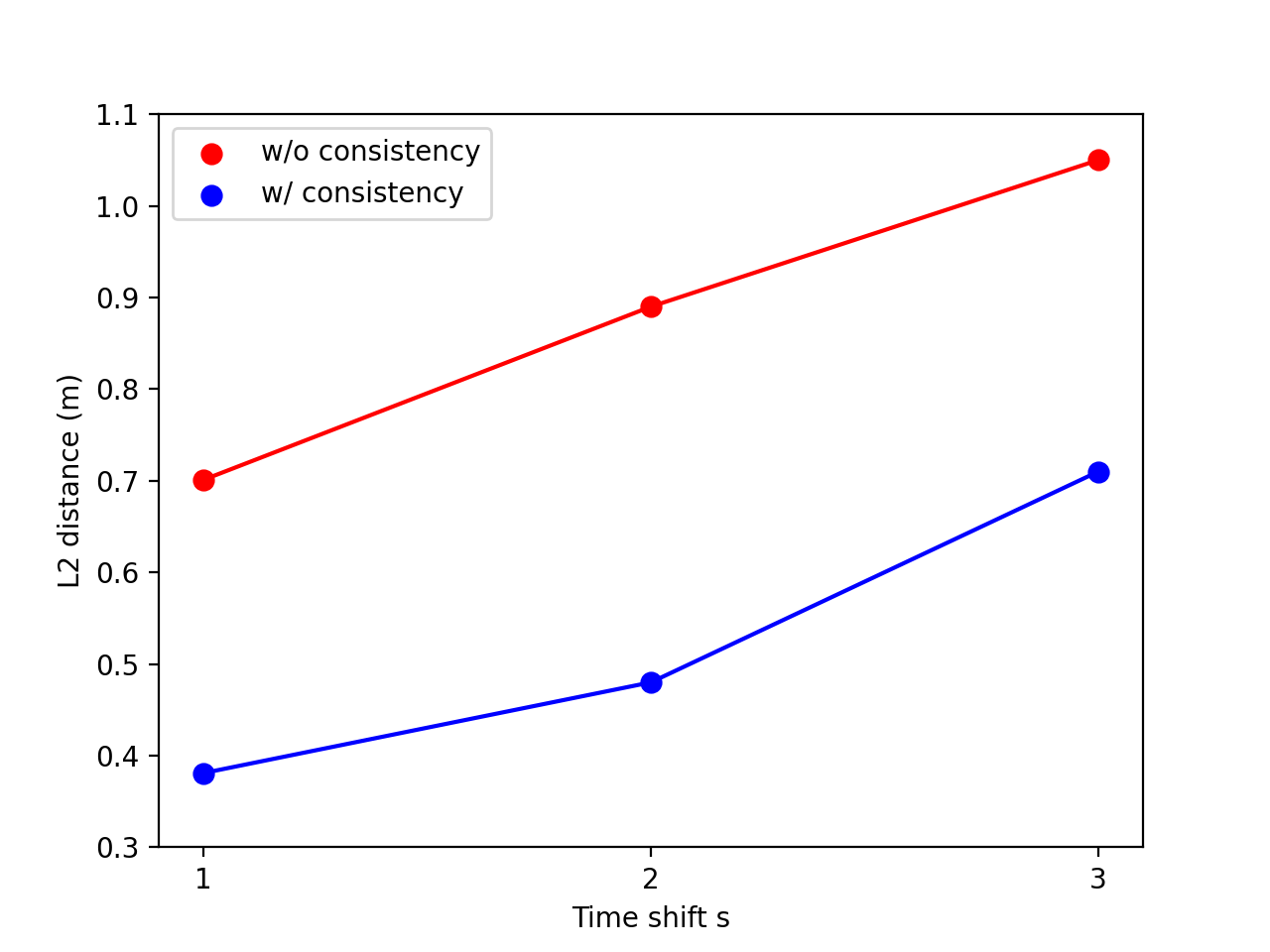}
    \caption{The L2 distance which varies with the time shift $s$.}
    \label{fig:consistency}
\end{figure}

\noindent\textbf{Reduction on Temporal Inconsistency.} We use the average L2 distance among all predicted trajectory waypoints to measure the temporal consistency. As shown in Fig.~\ref{fig:consistency}, our model without temporal consistency will have large inconsistency even though the time shift $s$ is small, which may lead to unstable behavior for the downstream task such as planning. With temporal consistency constraints, there is a significant improvement in the L2 distance divergence, demonstrating the effectiveness of our method. 

\noindent\textbf{Number of Teacher Targets.} As shown in Tab.~\ref{tab:Pseudo_Target_Num}, more teacher targets could bring better performance. Compared with $J=1$, $6$ teacher targets bring an extra nearly $1\%$ improvements. However, the marginal improvement decreases significantly so we finally choose $J=6$.

% Meanwhile, we also conduct experiments to find the best time-shift value $s$ in the temporal consistency. The details can be found in supplementary materials.
% As shown in Tab.~\ref{tab:time_shift}, choosing time shift $s=1$ has already achieved decent performance, with five out of six metrics ranking the first. Further increasing the $s$ will not bring much performance gain since the driving behavior could change a lot with large $s$.

% \vspace{-10px}
\subsection{Generalization Capability}
To verify the generalization capability of Dual Consistency Constraints and Teacher Targets Constraints, we also apply them to different models with state-of-the-art performance to show that they can be plugin-in training schemes.

\noindent\textbf{Consistency Component.} 
% Since our consistency constraints module is a general training strategy, we also apply it to different popular models with state-of-the-art performance to show its generalization capability. 
As shown in Tab.~\ref{tab:consistency_study}, our dual consistency constraints can effectively improve the performance of models regardless of their representations through the training phase. There is a noticeable improvement of over $5\%$ on every metric, especially for minFDE.

% \begin{table*}[t]
% \caption{Ablation study of Teacher Target Constraints (TTC) on different state-of-the-art models}
% \setlength\tabcolsep{3pt}
% % \vspace{-5px}
% 	\begin{center}
%          \scriptsize
% 		\begin{tabular}{c|c|cc|cc}
% 			\hline
% 			\multirow{2}* {Method} &
% 			\multirow{2}* {TTC} &
% 			\multicolumn{2}{c|}{K=1} & \multicolumn{2}{c}{K=6}\\
% % 			\cline{1-3}
% 			&  & minADE & minFDE & minADE & minFDE\\
% 			\hline 
% 			\multirow{2}{*}{LaneGCN~\cite{liang2020learning}} &$\times$ & 1.35&2.97 &0.71 &1.08 \\
% 			&\checkmark  &\textbf{1.30}  &\textbf{2.88} &\textbf{0.69} & \textbf{1.04} \\
% 			\hline
% 			\multirow{2}{*}{TPCN~\cite{ye2021tpcn}} &$\times$ & 1.34&2.95 &0.73 &1.15 \\
% 			&\checkmark  &\textbf{1.30} &\textbf{2.86} &\textbf{0.69} & \textbf{1.09} \\
% 			\hline
% 			%TODO(need to refine)%
% 			\multirow{2}{*}{mmTransformer~\cite{liu2021multimodal}} &$\times$ & 1.38&3.03 &0.71 &1.15 \\
% 			&\checkmark  &\textbf{1.29} &\textbf{2.80} &\textbf{0.68} & \textbf{1.04} \\
% 			\hline
% 			\multirow{2}{*}{DenseTNT~\cite{gu2021densetnt}} &$\times$ & 1.36&2.94 &0.73 &1.05 \\
% 			&\checkmark  &\textbf{1.30} &\textbf{2.82} &\textbf{0.69} & \textbf{1.00} \\
% 			\hline
% 		\end{tabular}
% 	\end{center}
% 	\vspace{-10px}
	
% 	\label{tab:mpt_study}
% % 	\vspace{-20px}
% \end{table*}

\noindent\textbf{Teacher Target.} Teacher-Target Constraints is another general training trick that can be widely used in other frameworks. In Tab.~\ref{tab:consistency_study}, we also verify its effectiveness on other public methods. Methods with Teacher-Target Constraints have nearly over $3$\% improvement in all metrics. For the original DenseTNT~\cite{gu2021densetnt}, we replace its original handcrafted optimization for teacher goal targets with our self-ensembling teacher targets. This strategy brings an over $5$\%  increase in performance, demonstrating the better quality of the self-ensembling teacher targets than handcrafted optimizations and estimation.

% \section{Limitations}
% Although the proposed method achieves state-of-the-art performance and can be universal components for motion forecasting tasks, it still has several drawbacks: 1). Our method lacks strong mathematical proof to support the proposed components; 2). It requires more training resources including GPUs to generate pseudo targets by self-ensembling. Meanwhile, when training is equipped with temporal consistency, the training time and memory usage will nearly double. 

\section{Conclusion}

In this work, we propose MISC, an effective architecture for the motion forecasting task. We impose inner-model dual consistency regularization on both spatial and temporal domains to leverage the potential of self-supervision, which has been ignored by previous efforts. Besides, we explicitly model the multi-modality by providing supervision and constraints with powerful self-ensembling techniques in an intra-model aspect. 
Experimental results on the Argoverse motion forecasting dataset and Waymo dataset show the effectiveness of our approach and generalization capability to other methods.

\clearpage

{\small
\bibliographystyle{ieee_fullname}
\bibliography{egbib}
}

\end{document}